\documentclass{article}

\usepackage[final, preprint]{neurips_2026}
\usepackage{amsmath}
\usepackage{multirow}
\usepackage[separate-uncertainty]{siunitx}
\usepackage{wrapfig} 
\usepackage{caption} 
\usepackage[dvipsnames]{xcolor}

\DeclareMathOperator{\sol}{\mathbf{u}}

\DeclareMathOperator{\bfx}{\mathbf{x}}

\DeclareMathOperator{\teacher}{\mathcal{T}}
\DeclareMathOperator{\student}{\mathcal{S}}

\usepackage{array}
\usepackage{booktabs}
\usepackage{pifont}

\newcommand{\meanstd}[2]{#1{\tiny\,$\pm$ #2}}
\usepackage{placeins}

\newcolumntype{C}[1]{>{\centering\arraybackslash}p{#1}}

\usepackage[utf8]{inputenc} %
\usepackage[T1]{fontenc}    %
\usepackage[hidelinks]{hyperref}       %
\usepackage{url}            %
\usepackage{booktabs}       %
\usepackage{amsfonts}       %
\usepackage{nicefrac}       %
\usepackage{microtype}      %
\usepackage{xcolor}         %

\usepackage{amsmath}
\usepackage{amssymb}
\usepackage{mathtools}
\usepackage{amsthm}
\usepackage[capitalize]{cleveref}
\Crefname{section}{Sec.}{Secs.}
\Crefname{figure}{Fig.}{Figs.}
\Crefname{table}{Tab.}{Tabs.}

\title{Distillation of Foundation Models for Time-dependent PDEs}

\author{
    \textbf{Daniel Musekamp\textsuperscript{1}}\thanks{Corresponding author: \href{mailto:daniel.musekamp@ki.uni-stuttgart.de}{daniel.musekamp@ki.uni-stuttgart.de}} \quad 
    \quad \textbf{Boshra Ariguib}\textsuperscript{1, 2} \quad
    \quad \textbf{Andrei Manolache}\textsuperscript{1,3} \quad \textbf{Mathias Niepert}\textsuperscript{1} \\
    \\
    \textsuperscript{1}University of Stuttgart \quad 
   \textsuperscript{2}{University of Cologne} \quad 
    \textsuperscript{3}{Bitdefender} \\ 
}

\begin{document}

\maketitle

\begin{abstract}
Foundation models for time-dependent partial differential equations (PDEs) are trained on large and diverse collections of physical systems and can generalize effectively to new downstream tasks. After fine-tuning on only a few trajectories from a target domain, they can achieve strong accuracy in low-data regimes. However, these models are typically large and computationally intensive, limiting their usefulness as fast surrogates for numerical solvers. 
We propose Teacher Rollout Extension (TREX), a knowledge distillation framework that transfers the predictive capability of a pretrained foundation model into a compact and efficient student. Starting from a fine-tuned teacher, TREX augments limited downstream data by generating long synthetic trajectories through teacher rollouts, optionally with periodic noise injection. This procedure samples from the teacher-induced rollout distribution without requiring explicit knowledge of the initial-condition distribution, while exposing the student to long-horizon states and local recovery behavior around states encountered during autoregressive prediction. The student can further incorporate task-specific inductive biases, such as equivariance, that the teacher does not necessarily enforce.
We evaluate TREX on multiple PDE benchmarks. The resulting students can match or surpass the teacher’s accuracy while reducing the number of parameters by several orders of magnitude and achieving more than an order-of-magnitude speedup in inference.
\end{abstract}

\section{Introduction}
Spatiotemporal physical processes such as fluid dynamics or cell growth can be described using partial differential equations (PDEs). Hence, solving PDEs is integral to science and engineering, for example, in astrophysics, design optimization, or weather prediction. Since solving PDEs for large systems and nonlinear dynamics using numerical solvers can be slow, recent research has investigated applying neural networks as surrogate models to learn solutions to PDEs. Compared with classical solvers, such approaches are differentiable out of the box and can incorporate experimental data, thereby enabling the modeling of effects not included in the simulation model. One core problem with neural PDE surrogates is that they require a training dataset, which is often generated using an expensive numerical solver, thereby making the surrogate modeling potentially uneconomical. 

As a remedy, foundation models for PDEs have been introduced \citep{mpp, herde2024poseidonefficientfoundationmodels, mccabe2025walrus}. By pretraining a surrogate on a large set of diverse physics, foundation models can transfer knowledge to  downstream tasks with fewer training samples. On the downside, their inference is computationally expensive, as they require more parameters to generalize well. 
Slow inference prohibits their use in applications that have real-time requirements, such as planning and control tasks in manufacturing or energy systems \citep{tian2025physics,janssens2024towards}. Furthermore, the used architectures are mostly large transformers that lack inductive biases, since these may vary from problem to problem. For example, many PDEs are locally equivariant to rotations and translations, and globally as well if the boundary conditions are periodic. Including these equivariances in the architecture can make models more effective and data-efficient \citep{zhdanov2024clifford, helwig2023group}.

\begin{figure}
    \centering
    \includegraphics[width=1\linewidth]{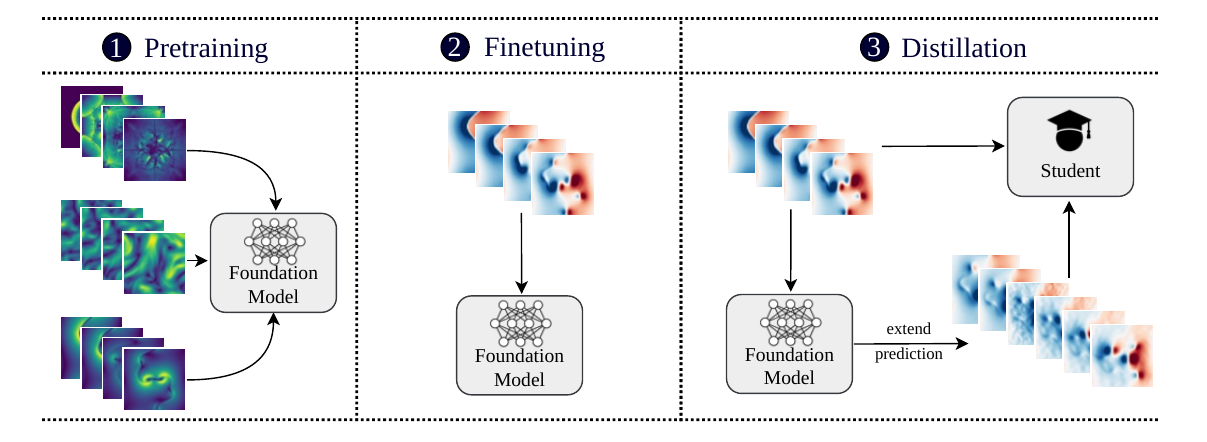}
    \caption{Foundation models are pretrained on a variety of PDE prediction tasks first (1), enabling them to adapt to a new task with only a few trajectories (2). To overcome their high inference time, we propose to distill the foundation model into a smaller and potentially equivariant student by extending the training trajectories using noisy rollouts of the teacher (3).}
    \label{fig:method}
\end{figure}

Therefore, we propose using knowledge distillation (KD) to reconcile the trade-off between fast inference and strong generalization (\cref{fig:method}).  Specifically, the pretrained foundation model is first fine-tuned on a small set of trajectories in the target domain. Afterward, the foundation model is used as the teacher to amend the available data with additional information. To our knowledge, we present the first systematic study of distilling general pretrained PDE foundation models into compact downstream surrogate models. Most prior work relies on specialized architectures or assumes full knowledge of the initial state distribution \citep{zhang2025omnifluids, li2025frequency}. 
However, access to the downstream initial-condition distribution is not always available. This is particularly relevant when data originates from experiments or partially observed systems, where only a limited set of trajectories is given. 

Hence, we introduce Teacher Rollout Extension (\textbf{TREX}), a distillation technique that avoids the need to sample new initial conditions. Starting from the few available downstream trajectories, TREX generates long teacher rollouts and trains the student on the resulting teacher-labeled trajectories. This exposes the student to the teacher-induced rollout distribution, i.e., the distribution of states encountered during autoregressive prediction, rather than only the short trajectories observed in the downstream data. Periodic noise injection can further broaden local state-space coverage and
teach the student how the teacher evolves from perturbed states. To transfer the teacher’s knowledge to the student model, we add these teacher-generated trajectories to the student dataset and train on a mixture of ground-truth and teacher-labeled transitions. 

To overcome the aforementioned lack of inductive biases in foundation models, we propose distilling the teacher into a student with task-appropriate inductive biases. For example, we use a tensorized FNO \citep{tfno} as the student. In addition to being equivariant to input translations, FNOs also exhibit inductive biases such as excluding high-frequency modes that are prone to noise. 

We investigate the effectiveness of our method through experiments with the Poseidon \citep{herde2024poseidonefficientfoundationmodels} and Walrus \citep{mccabe2025walrus} foundation models across a range of fluid dynamics tasks, demonstrating that TREX can achieve teacher accuracy while drastically reducing parameter count and inference time. Overall, our work provides the following contributions:
\begin{enumerate}
    \item To our knowledge, we present the first systematic study of distilling general pretrained PDE foundation models into compact downstream surrogate models.
    \item 
    We introduce TREX, a novel distillation method that generates supervision data using noisy teacher rollouts, in some cases surpassing the teacher's accuracy.
    \item We show that by distilling the foundation model into a small, lightweight student architecture with suitable inductive biases, we can reduce inference time by more than an order of magnitude while respecting physical properties such as equivariances.
\end{enumerate}
\section{Problem Definition}
We consider time-dependent PDEs on a spatial domain $\mathcal{X}$ and time
interval $[0,T]$, with solution
$\sol:[0,T]\times\mathcal{X}\to\mathbb{R}^{N_c}$:
\begin{align}
\partial_t \sol
&= F\left(t,\bfx,\sol,\partial_{\bfx}\sol,
\partial_{\bfx\bfx}\sol,\ldots\right),
&& (t,\bfx)\in[0,T]\times\mathcal{X}, \label{eq:pde}\\
\sol(0,\bfx)
&= \sol^0(\bfx),
&& \bfx\in\mathcal{X}, \label{eq:ic}\\
\mathcal{B}[\sol](t,\bfx)
&=0,
&& (t,\bfx)\in[0,T]\times\partial\mathcal{X}. \label{eq:bound}
\end{align}
Here, $F$ defines the dynamics, 
$\sol^0$ is the initial condition, and $\mathcal{B}$ specifies the boundary
condition. A task corresponds to solving this initial-value problem for
$\sol^0\sim p_0$. For learning, we use discretized solution trajectories
$(\sol^0,\sol^1,\ldots,\sol^{N_T-1})$, where each state
$\sol^i\in\mathbb{R}^{N_x\times N_c}$ contains the solution on $N_x$ spatial grid
points and $N_c$ channels. A neural PDE surrogate then learns a discrete
time-stepper $\mathcal{G}$ and is rolled out autoregressively $\sol^{i+1}\approx \mathcal{G}(\sol^i)$.  For models that require multiple context frames, we treat the complete context
window as the autoregressive state. Hence, each transition shifts this context by one frame and appends the newly predicted state.

\section{Related Work} \label{sec:pdefm}

Knowledge distillation \citep{hinton2015distilling, gou2021knowledge} aims to provide the student with additional information from the teacher network. For classification tasks, the class logits can be transferred to the student \citep{ba2014deep,hinton2015distilling}. For general tasks, the student can be trained to have feature representations close to those of the teacher \citep{romerofitnets}. Relational KD \citep{park2019relational} trains the student by comparing the relationship between the student's feature representations of different inputs and the teacher's, i.e., pushing the student to have the same distance between inputs as the teacher. Thus, the teacher and student do not need the same feature space representation. Similarly, \citet{lin2025perspective} introduce a region-aware attention mechanism that reconstructs student patches into representations more closely aligned with the teacher’s view, even when the two models have substantially different architectures. Lastly, the student can also be trained to match the teacher's derivatives. \cite{srinivas2018knowledge}  train a student by matching the teacher's Jacobian matrix. \cite{amintowards} apply a similar approach to distill a foundation model for molecules into a smaller student model by matching the teacher's energy-prediction Hessian. 
In addition to providing additional information about the encountered training examples, it is also possible to artificially generate new training instances using the teacher model. For example, \citet{lopes2017data} introduce data-free distillation, which aims to generate new samples that match the statistics of a dataset for which only those statistics are known. 

Distillation has also been applied to PDEs or spatiotemporal predictions in general. \cite{li2025frequency} distill a special teacher architecture with branches for low- and high-frequency features into the student using separate losses for each branch. \citet{wan2025sino} propose SINO, which first learns a fine-step surrogate from a small number of trajectories and subsequently distills it into a coarse-step FNO using a larger teacher-generated dataset. The synthetic trajectories are generated from newly sampled initial conditions, whereas TREX targets settings in which no such initial-condition sampler is available and instead expands the available downstream trajectories through teacher rollouts. \cite{zhang2025omnifluids} present Omnifluids. Omnifluids first trains a high-resolution model on a PDE with varying parameters using a finite-difference loss for a known equation, similar to PINO \citep{li2024physics}. The teacher is distilled into a lower-resolution model with the same architecture by matching the teacher's predictions on samples from the IC, assuming the teacher's parameter distribution is known. In contrast, we consider general foundation models pretrained on large datasets, and do not assume knowledge of the PDE or input distributions.

\paragraph{PDE Foundation Models.} In recent years, a number of PDE foundation models have been presented. Most use large vision transformers as their backbone \citep{mpp, liu2024prose, herde2024poseidonefficientfoundationmodels, mccabe2025walrus}. The earliest example of a PDE foundation model for nonlinear problems is MPP \citep{mpp}, which also uses axial attention to reduce the computational load. Since then, model and dataset sizes have been increasing. Poseidon \citep{herde2024poseidonefficientfoundationmodels} is based on a SwinV2 \citep{liu2022swin} vision transformer trained to predict the future state with the step size as a model input. Hence, it can be employed both for direct and autoregressive prediction. Walrus \citep{mccabe2025walrus} is trained on The Well data \citep{ohana2024well} and can solve 2D and 3D tasks. While using a vision transformer, it also takes multiple time steps as inputs to infer the underlying physics from the observed trajectory segment. \citet{Unisolver} present Unisolver, which provides the vision transformer backbone with additional information about the PDE, such as the domain geometry or the physical parameters. Most notably, the equation symbols are embedded using an LLM.
In contrast, PDEformer-2 \citep{ye2025pdeformer} uses a graph-based transformer, and DPOT \citep{hao2024dpot} is based on  Fourier attention layers. \cite{ups} propose to use a pretrained LLM backbone to propagate in time. Additionally, they also embed additional inputs using an LLM similar to Unisolver. For more related work, see \cref{related_work-appendix}.

\section{Knowledge Distillation of PDE Foundation Models}
We are given a small training set of trajectories $\sol$ from a downstream target domain 
\begin{equation}
\mathcal{D}_{\mathrm{GT}}
=
\{(\sol_n^0,\ldots,\sol_n^{N_{T}-1})\}_{n=1}^{N},
  \end{equation}
where each trajectory consists of $N_T$ states. Our goal is to build an accurate and efficient surrogate model by infusing knowledge from a foundation model (FM) used as the teacher $\teacher$, that autoregressively predicts the next state $\mathbf{z}^{t+1} = \teacher(\mathbf{z}^{t})$. Since the FM was pretrained on a large set of PDE problems, it can provide a lower error on the downstream task than a model trained from scratch. In the first step, the FM is fine-tuned on the available downstream task data. Afterward, we distill the teacher's knowledge into the student model $\student$. A natural distillation strategy is to sample new initial conditions 
$\sol_0 \sim p_0$, roll out the teacher, and train the student on the resulting teacher-labeled pairs \citep{zhang2025omnifluids}. 
However, in downstream applications, the initial condition distribution $p_0$
may be unknown, for example, for data coming from experimental sources.  In other cases, it may be difficult to sample directly from this distribution. Lastly, for FMs that require multiple context frames, sampling an initial
condition alone is insufficient to initialize the teacher, since generating
a valid context requires additional simulator calls.

\subsection{Knowledge Distillation using Teacher Rollout Extension}

We propose \emph{Teacher Rollout Extension} (TREX), a distillation procedure that trains the student not only on the available ground truth downstream trajectories but also on the state distribution induced by the fine-tuned teacher during autoregressive prediction. This distinction is important for PDE surrogates: at test time, the model is repeatedly applied to its own previous predictions, and therefore encounters states that need not lie exactly on the finite set of observed trajectories. 

TREX targets this autoregressive state distribution directly. It generates additional training trajectories by rolling out the teacher from the available ground truth states.
For simplicity, we state the following occupancy measure for the single-frame case used in our main Poseidon experiments. For multi-frame models, the same construction applies by replacing each state with its corresponding context
window. Let $\delta_{\sol_n^0}$ be the Dirac measure concentrated at $\sol_n^0$, and let 
\begin{equation} 
\nu_0
=
\frac{1}{N}\sum_{n=1}^N \delta_{\sol_n^0}
\end{equation}
be the empirical distribution over the available downstream initial states. A standard one-step training objective only supervises the model on transitions drawn from the short observed ground truth trajectories. In contrast, TREX constructs an extended teacher-induced state occupancy measure
\[
\rho_{\teacher}^{K}
=
\frac{1}{K}\sum_{t=0}^{K-1}(\teacher^t)_{\#}\nu_0,
\]
where $(\teacher^t)_{\#}\nu_0$ denotes the pushforward of $\nu_0$ under
$t$ autoregressive teacher steps and $K$ is the rollout horizon.
Hence, the occupancy measure describes the state distribution induced by
rolling out the teacher from the selected set of ground-truth samples.
This deterministic occupancy measure motivates TREX by describing the
additional states made available through teacher rollout extension.
In practice, TREX further perturbs this rollout distribution through
periodic noise injection, as described below.

\begin{figure}[t]
    \centering
    \includegraphics[width=1\linewidth]{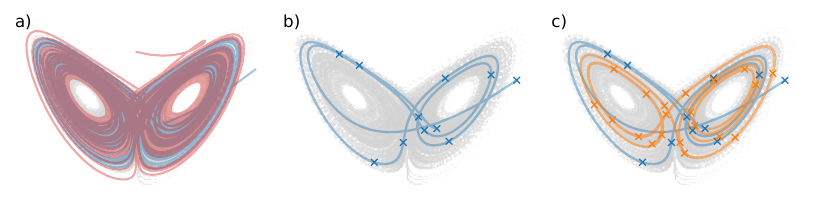}
    \caption{Long rollouts can increase state-space coverage in dynamical systems. Two trajectories of the Lorenz system \citep{lorenz1963deterministic} visit similar regions of state space over time (a). Starting from a short observed trajectory (b), extending the rollout generates additional states from the long-time trajectory distribution (c). 
    }
    \label{fig:lorenz}
\end{figure}

Thus, TREX does not require access to the true initial-condition distribution. Instead, it reuses the few ground-truth trajectories as anchors and expands them through the dynamics learned by the foundation model. This perspective connects TREX to the long-time behavior of dynamical systems. Many dissipative PDEs and chaotic systems possess recurrent long-time structure, such as attractors or invariant measures, so that trajectories initialized from different states eventually visit similar regions of the state space. If the teacher dynamics admit an invariant measure $\mu_{\teacher}$ satisfying
$\teacher_{\#}\mu_{\teacher}=\mu_{\teacher}$, then, under additional ergodicity
or mixing assumptions, sufficiently long teacher rollouts may approach this
long-time distribution. More generally, rollout extension increases coverage of states reachable under the teacher dynamics from the available downstream trajectories. The ergodic Lorenz attractor example in \cref{fig:lorenz} illustrates this effect: although trajectories start from different initial conditions, their long rollouts repeatedly visit a shared region of the state space. TREX exploits the same principle for PDE foundation models by turning a small number of ground-truth trajectories into a much broader set of teacher-labeled states.

\subsection{Noisy Teacher Rollouts}

Deterministic rollouts sample states on the teacher trajectory, but may provide
limited coverage of nearby off-trajectory states. Additionally, it can be
beneficial to include data points near typical rollout states in order to teach
the student how the dynamics evolve from perturbed inputs. Hence, when
appropriate, we inject noise periodically during the teacher rollout:
\begin{equation}
\tilde{\mathbf{z}}^t
=
\mathbf{z}^t + m_t \boldsymbol{\xi}^t,
\qquad
\mathbf{z}^{t+1}
=
\teacher\left(\tilde{\mathbf{z}}^t\right),
\qquad
m_t = \mathbf{1}\{t > 0 \;\land\; t \equiv 0 \mod T_{\mathrm{Noise}}\},
\end{equation}
\begin{equation}
(\boldsymbol{\xi}^t)_c(\mathbf{x})
\sim
\mathcal{N}(0,\sigma^2 s_c^2).
\end{equation}
Here, $\boldsymbol{\xi}^t$ is Gaussian noise applied independently at each
spatial location and channel, $s_c$ denotes the global standard deviation
of channel $c$ computed over the dataset, and $\sigma$ is a fixed scaling
factor. The noise is not applied at every step, but at intervals of
length $T_{\mathrm{Noise}}$, so that we obtain unperturbed sub-trajectories for
training and allow the teacher rollout to return toward states typical of its
unperturbed rollout distribution after a perturbation. Adding noise during
training is a common strategy for autoregressive neural PDE
solvers~\citep{sanchez2018graph, pfaff2020learning,takamoto2023learning}.
In these approaches, the training inputs may be perturbed while the target
trajectory remains unchanged. In TREX, by contrast, the teacher is explicitly
evaluated after perturbation, so that the synthetic target reflects the teacher
dynamics starting from the perturbed state.

By performing noisy rollouts from the initial conditions of the available
ground-truth trajectories, we obtain the dataset
$\mathcal{D}_{\mathrm{TREX}}$. It consists of teacher-generated
sub-trajectories between consecutive noise injections. A segment starting
at a noise-injection step $t_i$ is of the form
\[
\left(
\tilde{\mathbf z}^{t_i},
\mathbf z^{t_i+1},
\ldots,
\mathbf z^{t_i+T_{\mathrm{Noise}}}
\right).
\]
These segments are subsequently split into sub-trajectories of the desired
student rollout length.
 
The final student objective combines ground-truth supervision with the teacher
rollout distillation:
\[
\mathcal{L}_{\student}
=
\mathcal{L}_{\mathrm{GT}}
+
\lambda_{\mathrm{TREX}}
\mathcal{L}_{\mathrm{TREX}}.
\]
For both ground-truth and teacher-generated sub-trajectories, we train the
student using a $k$-step autoregressive rollout loss. For TREX data,
\[
\mathcal{L}_{\mathrm{TREX}}
=
\frac{1}{|\mathcal{D}_{\mathrm{TREX}}|} \sum_{\left(\mathbf{z}^i, ..., \mathbf{z}^{i+k}
\right) \in \mathcal{D}_{\mathrm{TREX}}} \frac{1}{k}\sum_{t=i}^{i+k - 1}
\ell\left(\mathbf{z}^{t+1}, \student (\hat{\sol}^t) \right), \quad \hat{\sol}^i = \mathbf{z}^i,  \quad \hat{\sol}^{t+1} = \student(\hat{\sol}^{t}).
\]
The training sub-trajectories
$\left(\mathbf{z}^i,\ldots,\mathbf{z}^{i+k}\right)$
are obtained by splitting the noise-free teacher-generated segments into
sequences of the desired rollout length. $\mathcal{L}_{\mathrm{GT}}$ is defined analogously on sub-trajectories extracted from the available ground-truth trajectories.
As TREX merely provides additional data points, it could be combined with other methods, such as feature-based distillation techniques, in the future.

\subsection{Enforcing Physical Constraints}
\label{sec:physical-constraints}

Foundation models for PDEs are usually constructed using large transformer architectures (\cref{sec:pdefm}). This makes them flexible enough to model many different systems, but it also means that they often do not explicitly encode the symmetries of a specific downstream PDE. Once the downstream task is fixed, however, such a structure may be known and can be built into the student. We focus on equivariance as one such constraint. Let $G$ be a group of
transformations acting on the state space, and let $\tau_g u$ denote the
transformed state for $g\in G$. A model $\mathcal{M}$ is equivariant to $G$ if
\[
\mathcal{M}(\tau_g u) = \tau_g \mathcal{M}(u),
\qquad
\forall g\in G.
\]
For a PDE surrogate, this means that transforming the input state and then predicting the next state gives the same result as predicting first and then
transforming the output. Many physical systems satisfy such symmetries. For
example, on a periodic domain, the solution operator of a translation-invariant
PDE satisfies $\Phi_{\Delta t}(\tau_a u) = \tau_a \Phi_{\Delta t}(u),$
where $\tau_a$ denotes a spatial shift. This provides a complementary benefit to distillation. The teacher contributes knowledge acquired during large-scale pretraining, while the student can impose task-specific structure known for the  PDE. In low-data settings, this is particularly useful, as the student does not need to infer symmetry solely from the few available trajectories. The student can therefore learn from the teacher-generated data while restricting the learned transition map to functions that better match the downstream physics. Architectural constraints can also act as regularizers. For example, a student's architecture may suppress unresolved high-frequency components, enforce locality, preserve conservation laws, or respect boundary conditions. Such constraints can reduce noisy artifacts and improve rollout stability.%

\section{Experiments}

For our main experiments, we use the Poseidon-L model \citep{herde2024poseidonefficientfoundationmodels} as the teacher. For simplicity, we always use the autoregressive evaluation of Poseidon (instead of the lead-time prediction mode). The teacher is distilled into a lightweight TFNO model \citep{tfno, kossaifi2025librarylearningneuraloperators}, which represents FNO's \citep{fno} weight matrix using a Tucker factorization to reduce the number of parameters but also improve generalization through regularization effects. While the teacher is trained with the original n-to-n training protocol, the student is trained autoregressively with 2-step sub-trajectories. More model and training details are listed in \cref{sec:model_details}. 
For TREX, the teacher is rolled out for 100 steps starting from the initial condition, where noise is injected after every 10 steps. The relative noise standard deviation is set to $\sigma=1$. As distillation baselines, we compare the original Poseidon-L teacher and the student trained only with the available ground truth data. Additionally, we add relational KD \citep{park2019relational} as a feature-based KD method, using Gaussian sketching to reduce the dimensionality to make the pairwise distances computationally feasible. Lastly, we add a baseline with full knowledge of the true initial condition distribution (IC-KD) similar to \cite{zhang2025omnifluids}. We use a total batch size of 32 and adjust the number of epochs so that the total number of batch updates is equal between all methods (25k when using only ground truth). The training times are listed in the appendix (\cref{tab:training_time}). TREX training is only slightly slower than pure ground truth training (e.g., increased training time from 272 to 291 min for 32 trajectories). For more details on the KD methods, see \cref{sec:kd_details}.

As downstream tasks, we use compressible Euler and incompressible Navier-Stokes equations from \cite{herde2024poseidonefficientfoundationmodels}. \cref{sec:pde_details} lists the used PDEs in detail. Since our goal is specifically to study whether useful knowledge from a foundation model can be transferred into a compact student, we focus on spatiotemporal Poseidon tasks for which the fine-tuned teacher outperforms the TFNO trained only on the available downstream data, as determined on the validation set. The trajectories contain 8 frames each. We report the original median relative $L^1$ error from Poseidon \citep{herde2024poseidonefficientfoundationmodels}, which normalizes the $L^1$ loss for each quantity of interest (QOI) frame-wise:
\[
L_{\mathrm{ QOI}}^1(\sol, \hat{\sol})  =
\frac{
\sum_{\mathbf{x}} \sum_{c_{\mathrm{QOI}}}
\left|
\hat{\sol}(t_{\mathrm{last}}, \mathbf{x}, c_{\mathrm{QOI}})
-
\sol(t_{\mathrm{last}}, \mathbf{x}, c_{\mathrm{QOI}})
\right|
}{
\sum_{\mathbf{x}} \sum_{c_{\mathrm{QOI}}}
\left|
\sol(t_{\mathrm{last}}, \mathbf{x}, c_{\mathrm{QOI}})
\right|
}.
\]\

For completeness, we also provide the results over all time steps in \cref{tab:main-results-all}. For the incompressible Navier-Stokes tasks, the QOI is the velocity vector. For compressible Euler (CE-RPUI), the mean of the median of all QOI errors is reported. We report the mean across three seeds for all main results, using different training trajectories  for each seed as well. 

\subsection{Main Results}

\begin{figure}
    \centering
    \includegraphics[width=1\linewidth]{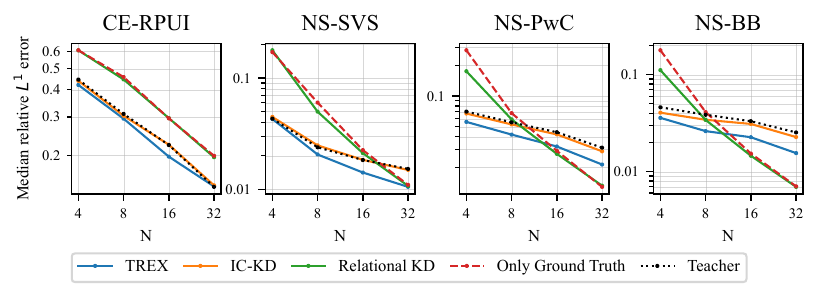}
    \caption{The median relative $L^1$ error for Poseidon downstream experiments. The distillation methods are compared to the teacher's accuracy over the number of trajectories in the training set N. TREX matches or surpasses the teacher across the selected downstream settings.} 
    \label{fig:main_res_poseidon}
\end{figure}
\begin{table}[t]
\small
  \centering
  \vspace{-1em}
  \caption{Inference time, GPU memory, and number of parameters for the Poseidon-L teacher and the TFNO student. Inference and memory are computed for a batch size of 64. }
  \label{tab:inference}
  \begin{tabular}{lccccc}
    \toprule
    \multirow{2}{*}{Model} & \multicolumn{2}{c}{1-step} & \multicolumn{2}{c}{Full rollout} & \multirow{2}{*}{\# Param.} \\
    \cmidrule(lr){2-3} \cmidrule(lr){4-5}
    & Time & Memory & Time & Memory & \\
    \midrule
    Poseidon-L (lead-time) & 270.9 ms & 3569 MB & 1908.2 ms & 10325 MB & 628.6M \\
    Poseidon-L (autoreg.) & 269.4 ms & 3605 MB & 1887.7 ms & 3749 MB & 628.6M \\
    TFNO (student) & 22.1 ms & 1059 MB & 152.1 ms & 1171 MB & 174.4K \\
    \bottomrule
  \end{tabular}
\end{table}

\cref{fig:main_res_poseidon} shows the main results for the Poseidon distillation. TREX matches or surpasses its teacher network across the selected low-data settings, which we will further investigate in the next section. IC-KD is also able to match the teacher on all investigated settings; however, it requires the full IC distribution to be known, which is not the case for experimental settings and might also be expensive (for example, for NS-BB, the IC is actually an already evolved state (\cref{sec:pde_details})). While relational KD can also slightly improve the student, we were not able to match the teacher in settings where the teacher outperformed ground truth learning. A reason might be the difficulty of meaningful distance measures in the large-dimensional feature fields encountered in the spatial neural PDE solver latent space.
Interestingly, we found that the TFNO is an exceptionally well-performing model for the considered tasks, even capable of outperforming the pretrained Poseidon in certain settings without KD.  In these cases, TREX can perform worse than only using the ground truth since the added teacher-generated data is moving the student away from the better solution it would have found using only ground truth. For the other student models considered below, we did not find similar results. Thus, distillation should only be applied if the foundation model was found to generalize well to the task at hand using a held-out validation set. \cref{fig:ablation}c) shows the error growth over time for the NS-PwC task, comparing the TFNO student with Poseidon. While the student has a larger error at the beginning, it has a more stable long-term rollout and has a slower error growth. Interestingly, this behavior is seen across multiple datasets (\cref{fig:all_error_growth}). While surpassing the teacher's accuracy, the student needs 3604 times fewer parameters (\cref{tab:inference}). The inference time is reduced by over an order of magnitude, and the memory consumption for a forward pass by a factor of 3.2 (compared to autoregressive Poseidon rollouts).

\subsection{Ablations}

To investigate why TREX can generate a student that outperforms its teacher, we conduct an ablation study. To this end, we select the NS-PwC-4 case and fully remove the ground-truth trajectories, training the student only on teacher-generated data (\cref{tab:ablation}). Even without the ground truth data, TREX still outperforms its teacher. Removing noise injections during the rollout reduces the gap, but not fully. The phenomenon that distillation can improve student performance has been observed in the general KD literature \citep{mobahi2020self, park2019relational}. It can be explained by regularization effects, which can even occur when distilling the same model into the same architecture \citep{zhang2019your, mobahi2020self}. 
\begin{figure}[t]
  \centering
    \includegraphics[width=0.94\linewidth]{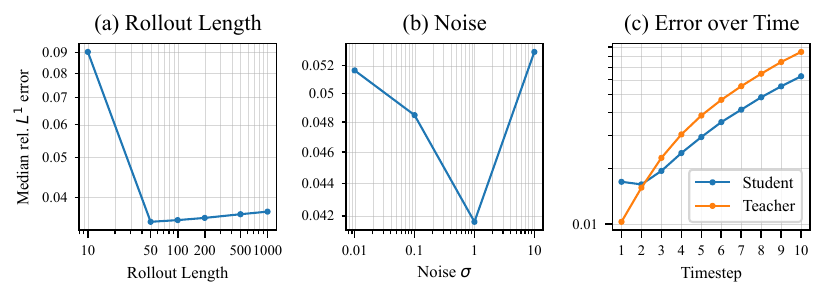}
    \captionof{figure}{Ablation study on NS-SVS (four trajectories). The effect of the rollout is relatively constant over a wide range of hyperparameters (a). Injecting noise also reduces the error, particularly at relatively high noise amplitudes (b). In (c), the error over time is shown for NS-PwC (8 trajectories). Time steps $t>7$ show temporal extrapolation beyond the training time horizon.  While the TREX-trained student has a higher one-step error than the teacher, the student has a more stable rollout.}
    \label{fig:ablation}
\end{figure}
\begin{wraptable}[15]{l}{0.44\linewidth}
\caption{Ablation study on NS-PwC (four trajectories). Shown are the effects of removing the noise injection during the teacher rollout, as well as removing ground truth trajectories fully from the student training. }
\begin{tabular}{lS[table-format=1.3(3)]}
        \toprule
        Variant & {Median rel. $L^1$ } \\
        \midrule
        TREX & 0.056(3) \\
        w/o GT & 0.060(2) \\ 
        w/o GT \& w/o Noise & 0.064(6) \\ 
        \midrule

        Teacher  & 0.070(11) \\ 
        \bottomrule
      \end{tabular}
       \label{tab:ablation}
\end{wraptable}
Next to the core algorithm choices, we also investigate the influence of the included hyperparameters. \cref{fig:ablation}a) shows the effect of the rollout length. The algorithm is quite robust to the choice of this hyperparameter. The performance only decreases slowly for very long rollouts. The effect of the strength of the injected noise is shown in \cref{fig:ablation}b). Interestingly, choosing the noise quite high at a level of one time the channel-wise standard deviation $s_c$ produces the best results.  \cref{fig:noisyrolloutsvs} shows an example of an extended noisy rollout (appendix). While the state appears very noisy after the noise injection, the teacher rollout effectively reduces the noise level in subsequent states. 
\begin{figure}[t]
    \centering
    \begin{minipage}[t]{0.41\linewidth}
                  \vspace{0pt}

        \centering
   
    \sisetup{table-format=1.3(3),detect-weight=true,detect-inline-weight=math}
    \centering
    \includegraphics[width=1\linewidth]{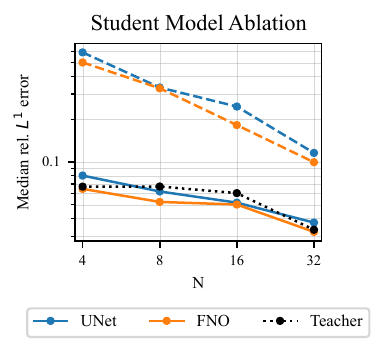}
    \caption{Ablation on the student architecture (NS-PwC). Shown are the results for a UNet and FNO trained with TREX (full) and only ground truth data (dashed).}
    \label{fig:modelabl}

    \end{minipage}
    \hfill
    \begin{minipage}[t]{0.55\linewidth}
            \vspace{0pt}

        \centering
        \includegraphics[width=1\linewidth]{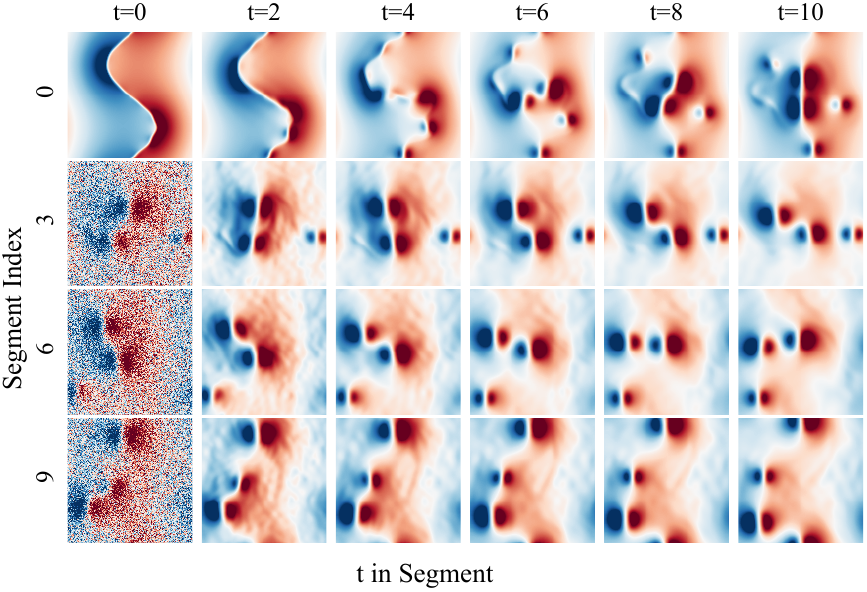}
        \caption{Example TREX rollout of the Poseidon teacher on the NS-SVS task ($v_x$ channel). Each row represents a segment of the complete rollout, starting with the state to which the noise was applied. }
        \label{fig:example_main}
    \end{minipage}
\end{figure}
We also provide an experiment with other student architectures. We use a modern UNet \citep{pdearena} and a standard FNO \citep{fno}. As shown in \cref{fig:modelabl}, TREX is also able to let other student architectures match the teacher's accuracy for all considered numbers of trajectories. While the TFNO was able to improve over the teacher using only ground truth for a higher number of trajectories in the main experiments (\cref{fig:main_res_poseidon}), this was not the case for the UNet or even a standard FNO, and was only possible through distillation. 
\begin{table*}[t]
    \centering
    \caption{Distillation of Walrus \citep{mccabe2025walrus} on the Kolmogorov Flow dataset. TREX improves long-rollout accuracy while reducing inference time by more than an order of magnitude.}
    \label{tab:teacher}
\begin{tabular}{lcccc}
\toprule
        & $\text{VRMSE}_{\text{One-Step}}$
        & $\text{VRMSE}_{T \in [1:20]}$
        & $\text{VRMSE}_{T \in [21 : 60]}$
        & {{Inference Time in s}} \\
\midrule
Walrus & $\mathbf{0.012}$ & 0.214 & 0.713 & 21.425 \\
Only GT & 0.050 & 0.355 & 0.972 & 0.401 \\
TREX & 0.016 & $\mathbf{0.123}$ & $\mathbf{0.6859}$ & 0.414 \\
\bottomrule
\end{tabular}
\end{table*}
\begin{table}[t]
\small
  \caption{
  Translation equivariance results. Left: median relative $L^1$ equivariance error, computed using the unperturbed predictions as ground truth. Right: final-step relative $L^1$ errors on original and shifted OOD rollouts.
  }
  \centering
  \setlength{\tabcolsep}{3pt}
  \begin{tabular}{lcccccc}
    \toprule
    & \multicolumn{2}{c}{Equivariance error}
    & \multicolumn{4}{c}{Final-step rollout error} \\
    \cmidrule(lr){2-3}
    \cmidrule(lr){4-7}
    Dataset
    & Teacher & Student
    & Poseidon & Poseidon shift
    & TFNO & TFNO shift \\
    \midrule
    NS-BB
    & \meanstd{0.026}{0.013} & \meanstd{0.000}{0.001}
    & \meanstd{0.0651}{0.0286} & \meanstd{0.0676}{0.0203}
    & \meanstd{0.0504}{0.0080} & \meanstd{0.0504}{0.0080} \\
    CE-RPUI
    & \meanstd{0.101}{0.006} & \meanstd{0.001}{0.001}
    & \meanstd{0.4189}{0.0767} & \meanstd{0.4108}{0.0856}
    & \meanstd{0.3549}{0.0344} & \meanstd{0.3549}{0.0344} \\
    NS-PwC
    & \meanstd{0.023}{0.006} & \meanstd{0.000}{0.001}
    & \meanstd{0.1095}{0.0420} & \meanstd{0.1041}{0.0385}
    & \meanstd{0.0744}{0.0206} & \meanstd{0.0744}{0.0206} \\
    NS-SVS
    & \meanstd{0.110}{0.005} & \meanstd{0.000}{0.001}
    & \meanstd{0.0820}{0.0083} & \meanstd{0.2398}{0.0086}
    & \meanstd{0.0755}{0.0072} & \meanstd{0.0755}{0.0072} \\
    \bottomrule
  \end{tabular}
  \label{tab:equivariance}
  \label{tab:transformed-metrics}
\end{table}

\subsubsection{Teacher Model}
We also experiment with other teacher models. Walrus \citep{mccabe2025walrus} is a larger vision transformer (1.3B parameters) that uses the last six states as context for next state prediction. This setup is crucially different since it does not allow IC-KD, as we would need to run the simulator to get the initial sequence to start the rollout. To also go beyond the rather short trajectories provided by the Poseidon datasets, we test distillation on the 2D Kolmogorov Flow dataset from \cite{li2022learning}  with a Reynolds number of 5000. The trajectories have 501 time steps, where the first 100 are discarded \citep{li2022learning}. For the student, we select a higher capacity TFNO with 64 modes and a feature dimension of 80, trained for only 10k epochs with batch size 64. Similar to the teacher, the student is using six input steps as the context. For the ground truth only baseline, we found a single step as input to work better.  We fine-tune Walrus on a single trajectory and report the VRMSE metric (\cref{eq:vrmse}) for different rollout intervals. The student trained with TREX can also successfully distill the multi-step teacher foundation model (\cref{tab:teacher}).  

\subsubsection{Equivariance}

To investigate the effect of the explicit equivariance built into the student architecture, we test how the models behave under spatial shifts on the periodic grid. We apply random 2D rolls to the input fields, which corresponds to the translation symmetry of the underlying periodic domain. An equivariant model should produce the same prediction up to this same roll. We evaluate this in two ways: first, as a direct equivariance error, and second, through shifted autoregressive rollouts. 

For the direct equivariance metric, let $a=(d_x,d_y)$ denote a sampled 2D periodic translation, and let $\tau_a$ be the corresponding shift operator. We compare the prediction on the shifted input with the shifted prediction on the original input, $\|f(\tau_a x)-\tau_a f(x)\|_1$, using the same median relative $L^1$ normalization as in the main evaluation. As shown in \cref{tab:equivariance}, the student has substantially lower equivariance error than the teacher across all datasets. This is expected from the student architecture, but it is still important: distillation transfers the teacher's predictive behavior into a model class that enforces a symmetry not explicitly respected by the teacher.
\begin{figure}[t]
    \centering
    \includegraphics[width=0.95\linewidth]{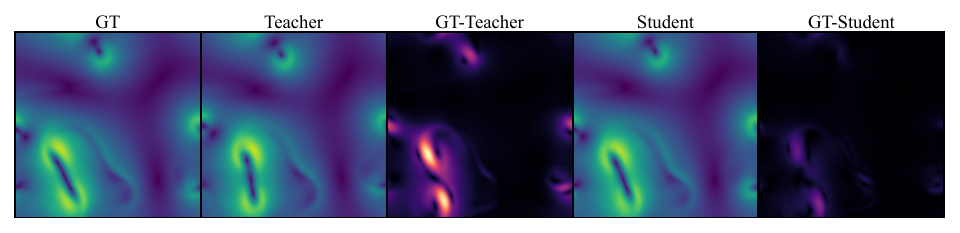}
    \caption{Final-step shifted-rollout comparison on NS-SVS. The teacher shows a visibly larger error with respect to the shifted ground truth (GT), while the student prediction remains much closer and the corresponding difference is substantially smaller.}
    \label{fig:equi-svs-reduced}
\end{figure}

We further evaluate the effect of equivariance in autoregressive rollouts for temporal extrapolation by applying the same spatial shift to the initial condition and comparing the final prediction to the shifted ground truth. The results in \cref{tab:equivariance} show that the shifted rollout error of the student remains essentially unchanged, while the teacher can degrade substantially, most clearly on NS-SVS. Since NS-SVS also has the largest teacher equivariance error, we show a qualitative example in~\cref{fig:equi-svs-reduced}. In this case, the teacher produces visibly worse shifted rollouts, while the student remains consistent with the translated dynamics. We include the entire rollout trajectory in Appendix~\cref{fig:apx-equi}. These results suggest that equivariance can be a useful property of the student, especially in autoregressive simulations where small symmetry errors may accumulate over long rollouts. More broadly, it is encouraging that TREX can use a non-equivariant teacher to train a compact student that preserves the teacher’s predictive behavior while enforcing the symmetry of the downstream task.

\section{Conclusion}
We introduce TREX, a simple and effective knowledge distillation framework for PDE foundation models. By extending limited downstream data through long and noisy teacher rollouts, TREX enables the student to match the teacher’s long-horizon dynamics, thereby providing valuable additional information. Our results demonstrate that distillation in the foundation model regime is not only feasible but highly effective: compact student models can match the accuracy of large teachers while achieving substantial improvements in inference speed and memory efficiency. Additionally, the student can incorporate inductive biases to faithfully reproduce prior knowledge about the system, such as equivariances. Overall, TREX bridges the gap between the strong generalization capabilities of large foundation models and the efficiency requirements of practical surrogate modeling. In the future, TREX could be combined with other distillation techniques to also provide more information on the influence of other factors such as physical parameters to the student.

\ack{
Funded by the German  Federal Ministry of Research, Technology and Space (BMFTR) as part of the InnoPhase project (funding code: 02NUK078). Additionally,  we acknowledge the support of Deutsche Forschungsgemeinschaft (DFG) under funding code 569019417 and the support of the Stuttgart Center for Simulation Science (SimTech). Andrei Manolache acknowledges funding by the EU Horizon project ELIAS (No. 101120237). The authors thank the International Max Planck Research School for Intelligent Systems (IMPRS-IS) for supporting Daniel Musekamp, Andrei Manolache, and Mathias Niepert. 
}

\bibliography{neurips.bib}
\bibliographystyle{neurips_2025}

\newpage
\appendix

\section{Limitations}
While TREX provides an effective framework for distilling PDE foundation models, it comes with several limitations. First, TREX relies on the quality of the fine-tuned teacher. Since the method samples from the teacher-induced rollout distribution, any systematic bias or instability in the teacher’s long-horizon dynamics may be transferred to the student. In particular, if the teacher converges to an incorrect or distorted attractor, TREX may reinforce this behavior. Although mixing teacher-generated data with ground-truth trajectories mitigates this issue, it does not fully eliminate it. Second, rollout extension is most useful when long trajectories continue to
explore informative regions of state space. Its benefit may therefore be
limited for systems that rapidly collapse to a trivial fixed point, exhibit
strongly transient dynamics, or contain multiple disconnected long-time
regimes that are not represented by the available trajectories.
Moreover, the Gaussian perturbations used by TREX are not guaranteed to
satisfy all physical constraints of the underlying PDE and should therefore
be viewed as a state-space augmentation strategy rather than as physically
valid perturbations.

\section{Broader Impact}

This work aims to advance machine learning methods for scientific simulation and time-dependent PDEs by making foundation-model-based surrogates more efficient and easier to deploy. Faster surrogate models could reduce the computational cost of simulation workflows and make such tools more accessible in scientific and engineering applications. We do not expect broad negative societal impacts from the method itself.

\section{Compute Resources}

The experiments were performed on a cluster with 10 NVIDIA GeForce RTX 4090 GPUs and on a larger compute cluster with 32 NVIDIA A100 GPUs and 32 NVIDIA L40 GPUs. Each individual training run used a single GPU. The reported inference-time and memory measurements for Poseidon experiments were collected for a batch size of 64, as described in the main text.

\section{Continued Related Work} \label{related_work-appendix}
Several authors have investigated explicitly including attractor dynamics for neural PDE solvers. \citet{li2022learning} include a loss term to enforce that the model always converges to the attractor on artificial data drawn uniformly from a fixed ball around zero. DySLIM \citep{schiffdyslim} enforces that the learned model follows the invariant measure by minimizing the Maximum Mean Discrepancy between the data distribution and the learned attractor. \cite{cheng2025learning} proposes a Poincaré Flow Neural Network, which is trained by dividing the data into the contracting and measure-invariant phases that can occur in a dissipative chaotic system. Depending on the phase, a loss is applied to enforce the contracting or invariant behavior.

\section{Datasets} \label{sec:pde_details}
\FloatBarrier

We mainly use the datasets provided by  \cite{herde2024poseidonefficientfoundationmodels}. All datasets, besides Kolmogorov Flow, contain 8 time steps, which end at $T=0.7$ and use periodic boundary conditions.  The  spatial resolution is 128x128 for a grid of $[0, 1]\times[0,1]$. The Poseidon problems have 120 validation and 240 test trajectories.

\subsection{NS-PwC}
The Navier-Stokes piecewise constants task uses the incompressible Navier-Stokes equations:
\begin{equation}
\partial_t \mathbf{u} + (\mathbf{u} \cdot \nabla)\mathbf{u} + \nabla p = \nu \Delta \mathbf{u}, \nabla \cdot \mathbf{u} = 0.
\end{equation}
For NS-PwC, the initial state is given as a set of constants on a grid of 10 cells in each dimension:
\begin{equation}
\omega_0(x,y)=c_{i,j}
\quad \text{for } (x,y)\in [x_{i-1},x_i]\times[y_{j-1},y_j],
\qquad i,j=1,\ldots,p.
\end{equation}
with $x_i = y_i = \frac{i}{p}, \, \text{for } i = 0,1,2,\dots,p$ and the constant square values drawn from $c_{i,j} \sim \mathcal{U}_{[-1,1]}$. Here, $\omega_0$ refers to the vorticity that can be transformed into the velocity in the x and y directions using the incompressibility condition.

\subsection{NS-BB}
In Fourier space, the Brownian bridge initial conditions are initialized as
\begin{equation}
    W(x) = \sum_{\lvert k \rvert_\infty \leq N} \frac{1}{\lVert k \rVert_2^{3/2}} \sum_{m,n,\ell \in \{0,1\}} \alpha_k^{(m n \ell)} \, \mathrm{sc}_m(x)\mathrm{sc}_n(x)\mathrm{sc}_\ell(x),
\end{equation}
where 
\begin{equation}
    \mathrm{sc}_i(x) =
\begin{cases}
\sin(x), & \text{for } i = 0, \\
\cos(x), & \text{for } i = 1
\end{cases}.
\end{equation}
W(x) does not directly give $u_0$. Instead, it defines the state at t=-0.5, and the initial conditions are derived by evolving the Navier-Stokes equations until t=0. The random variables are sampled from  $\alpha_k^{(m n \ell)} \sim \mathcal{U}_{[-1,1]}$.

\subsection{NS-SVS}
The next Navier-Stokes task uses Sinusoidal Vortex Sheet (SVS) initial conditions, which also use a vorticity formulation
\begin{equation}
    \omega_0^\rho = \phi_\rho * \omega_0,
\end{equation}
where  $\omega_0$ is defined as
\begin{equation}
\omega_0(x) = \delta(x - \Gamma) - \int_{\mathbb{T}^2} d\Gamma,
\end{equation}
and 
\begin{equation}
\phi_\rho(x) = \rho^{-2} \, \psi\!\left(\frac{\|x\|}{\rho}\right).
\end{equation}
$\psi(r)$ is defined as 
\begin{equation}
\psi(r) = \frac{80}{7\pi} \left[
(r+1)_+^3 - 4\left(r+\frac{1}{2}\right)_+^3
+ 6r_+^3 - 4\left(r-\frac{1}{2}\right)_+^3
+ (r-1)_+^3
\right],
\end{equation}
and $\Gamma$ as
\begin{equation}
\Gamma = \left\{ (x,y) \in \mathbb{T}^2 \;\middle|\;
y = \frac{1}{2} + 0.2\sin(2\pi x)
+ \sum_{i=1}^{p} \alpha_i \sin\big(2\pi(x + \beta_i)\big)
\right\}.
\end{equation}

The variable parameters are drawn from 
\begin{equation}
\alpha_i \sim \mathcal{U}_{[0,\,0.003125]}, \quad
\beta_i \sim \mathcal{U}_{[0,\,1]},
\end{equation}
and $\rho$ and $p$ are fixed to $\rho = \frac{5}{128}$ and $  p = 10$.

\subsection{CE-RPUI}
The Compressible Euler PDE is given as

\begin{equation}
u_t + \nabla \cdot F(u) = 0,
\end{equation}
where $u$ and $F(u)$ are defined as
\begin{equation}
u = \begin{bmatrix} \rho \\ \rho \mathbf{v} \\ E \end{bmatrix}, \quad F(u) =
\begin{bmatrix}
\rho \mathbf{v} \\
\rho \mathbf{v} \otimes \mathbf{v} + p \mathbf{I} \\
(E + p)\mathbf{v}
\end{bmatrix}.
\end{equation}
For the CE-RPUI problem, the four channels are drawn as constant values for each region of the input grid. The input division sets $D_{i,j}$ are computed using sinusoidal functions $\sigma_x(x,y)$ and $\sigma_y(x,y)$, which are defined as 
\begin{equation}
\sigma_x(x,y) = \sum_{i,j=1}^{p} \alpha_{x,i,j}
\sin\!\Big(2\pi\big(i + 2p^2\big)x + \big(j + 2p^2\big)y + \beta_{x,i,j}\Big),
\end{equation}
 and similarly
\begin{equation}
\sigma_y(x,y) = \sum_{i,j=1}^{p} \alpha_{y,i,j}
\sin\!\Big(2\pi\big(i + 2p^2\big)x + \big(j + 2p^2\big)y + \beta_{y,i,j}\Big).
\end{equation}
The coefficients $\alpha_{k,i,j}$  $\beta_{k,i,j}$  are drawn from uniform distributions:
\begin{equation}
\alpha_{k,i,j} \sim \mathcal{U}[-0.01,0.01], 
\qquad
\beta_{k,i,j} \sim \mathcal{U}[0,1].
\end{equation}
Afterward, the input space regions are defined as 
\begin{align}
D_{i,j} =
\left\{
(x,y) \in \mathbb{T}^2 \;\middle|\;
\begin{aligned}
&x_{\min} \le \{x + \sigma_x(x,y) +1\} < x_{\max}, \\
&y_{\min} \le \{y + \sigma_y(x,y) +1\} < y_{\max}
\end{aligned}
\right\},
\end{align}
where $\{x\} := x - \lfloor |x| \rfloor \, \mathrm{sgn}(x)$
 and
\begin{equation}
x_{\min} = \frac{i}{p+1}, \quad x_{\max} = \frac{i+1}{p+1}, \quad
y_{\min} = \frac{j}{p+1}, \quad y_{\max} = \frac{j+1}{p+1}.
\end{equation}
In the end, the initial condition is defined using the regions $D_{i,j}$
\begin{equation}
(\rho, v_x, v_y, p)\big|_{t=0} = (\rho_{i,j}, v^x_{i,j}, v^y_{i,j}, p_{i,j}) \quad \text{in } D_{i,j},
\end{equation}
where the region-wise parameters are drawn from
\begin{equation}
\rho_{i,j} \sim \mathcal{U}[1,3], \quad
v^x_{i,j} \sim \mathcal{U}[-10,10], \quad
v^y_{i,j} \sim \mathcal{U}[-10,10], \quad
p_{i,j} \sim \mathcal{U}[5,7].
\end{equation}

\subsection{Kolmogorov Flow}
For the ablations, we use the Kolmogorov flow dataset \citep{li2022learning} at a Reynolds number (Re) of 5000. The PDE can be described as 
\begin{align}
\frac{\partial u}{\partial t}
&=
-u\cdot\nabla u
-\nabla p
+\frac{1}{Re}\Delta u
+\sin(4y)\hat{x},
\qquad
&(x, y, t) \in  [0,2\pi]\times[0,2\pi]\times[0,\infty)\\
\nabla\cdot u&=0,
\qquad
&(x, y, t) \in  [0,2\pi]\times[0,2\pi]\times[0,\infty)\\
u(\cdot,0)&=u_0,
\qquad
&(x, y) \in  [0,2\pi]\times[0,2\pi].
\end{align}
We transform the PDE into velocity form so that it can be included in Walrus' channel mapping. The resolution is 128x128. The trajectories are 501 time steps long, where the first 100 are discarded. We use 48 trajectories for validation and test each.

\section{Model and Training Details} \label{sec:model_details}
All student models are trained to predict the difference, i.e.,
$\hat{\sol}^{t+1} = \student(\sol^{t}) = \mathcal{M}(\sol^t) + \sol^t$. The learning rate is set to $10^{-3}$ initially and scaled down to $10^{-6}$ using cosine annealing. We use the standard Adam optimizer \citep{adam} and a batch size of 32. The experiments were performed on NVIDIA GeForce RTX 4090 GPUs (one GPU per run). For the main results, we report the test error for the model checkpoint with the best error on the validation set for the ground-truth only and relational KD baselines, since we encountered overfitting for these methods. For TREX and IC-KD, we report the last model checkpoint to be fair to the Teacher baseline, which also did not use early-stopping. We use the original training loss from Poseidon over the batch $\mathcal{B}$:
\[
L_{\mathrm{train}}^{1}(\theta)
=
\frac{1}{N_{\mathrm{QOI}}}
\sum_{\mathrm{QOI}}
\frac{
\displaystyle
\sum_{b\in\mathcal{B}}
\sum_{t}
\sum_{\mathbf{x}}
\sum_{c_{\mathrm{QOI}}}
\left|
\hat{\sol}_{\theta,b}
\left(t,\mathbf{x},c_{\mathrm{QOI}}\right)
-
\sol_b
\left(t,\mathbf{x},c_{\mathrm{QOI}}\right)
\right|
}{
\displaystyle
\sum_{b\in\mathcal{B}}
\sum_{t}
\sum_{\mathbf{x}}
\sum_{c_{\mathrm{QOI}}}
\left|
\sol_b
\left(t,\mathbf{x},c_{\mathrm{QOI}}\right)
\right|
+
\varepsilon_{L_1}
}.
\]

\paragraph{TFNO.} We use the Tucker factorized FNO \citep{tfno} version with the implementation provided by \cite{kossaifi2025librarylearningneuraloperators}. 
 The factorization reduces the number of parameters with a compression factor of 0.1. The network uses 32 modes per dimension, 32 hidden channels, three layers, and a GELU~\citep{hendrycks2016gaussian} activation function. 

\paragraph{U-Net.} For the student model ablations, we use the modern U-Net \citep{ronneberger2015u} as provided by \cite{pdearena}. The U-Net has 4 levels of resolution, increasing the number of channels per level by 1, 2, 2, and 4, starting at 16 hidden channels after the embedding, resulting in 9,161,332 parameters. 
For U-Net, we found it advantageous to train with a 5-step error for the model ablation experiment.

\paragraph{FNO.}
We also use a traditional FNO \citep{fno} without factorization for the model ablations. The model uses the same number of modes and channels as the TFNO, resulting in 5,042,836 parameters. The FNO uses 4 layers.

\paragraph{Poseidon.}As the teacher, we use the Poseidon-L model \citep{herde2024poseidonefficientfoundationmodels}.  Poseidon-L was pretrained on six different datasets containing Navier-Stokes and compressible Euler equations with different sets of initial conditions. The model uses a SwinV2 transformer as its backbone, a multi-resolution vision transformer.  The architecture operates on a latent representation with embedding dimension \(C = 192\). For each hierarchical level \(i\), the model employs \(t_i = 8\)  blocks, resulting in a total of 629M trainable parameters. The network consists of \(L = 4\) levels, connected via \(L-1\) downsampling and upsampling operations to progressively refine spatial resolution. The number of attention heads per level is defined as:
\[
[h_1, h_2, h_3, h_4] = [3, 6, 12, 24],
\]
allowing increased representation capacity at coarser scales. In addition, each level incorporates two ConvNeXt layers \citep{convnet}. The model operates on input patches of size \(p = 4\) and a window size of \(M = 16\).

Training is performed using the AdamW optimizer \citep{adamw} with cosine learning rate decay. The main learning rate is initialized as \(\eta_b = 5 \cdot 10^{-5}\). A weight decay of \(10^{-6}\) is applied for regularization. The batch size is set to 16, and the model is trained for 200 epochs without early stopping. Gradient norms are clipped at a maximum value of 5 to ensure numerical stability.

\paragraph{Walrus.}
For Walrus \citep{mccabe2025walrus} fine-tuning, we used only 50k sample presentations to the model (batch size 4). We used global normalization and followed the original training procedure otherwise. \\
For  the distillation, we started 50 rollouts from different random sub-sequences in the trajectory of length 100. We found noise during Walrus rollout harmful, likely since adding noise to a multi-frame input does not just create a different state, but also suggests different physics to the model if it learned to infer it from its inputs. More importantly, it was crucial to deactivate the jitter added in Walrus inference, as this creates stochastic predictions and hence noisy labels. Walrus uses the VRMSE metric:
\begin{equation} \label{eq:vrmse}
\operatorname{VRMSE}_{t_1:t_2}
\bigl(\widehat{\mathbf{u}},\mathbf{u}\bigr)
=
\frac{1}{t_2-t_1+1}
\sum_{t=t_1}^{t_2}
\sqrt{
\frac{
\left\langle
\left|
\widehat{\mathbf{u}}_t-\mathbf{u}_t
\right|^2
\right\rangle
}{
\left\langle
\left|
\mathbf{u}_t-\overline{\mathbf{u}}_t
\right|^2
\right\rangle+\varepsilon
}
}.
\end{equation}
We use the normalize mean absolute error as the loss function
\[
\mathcal{L}_{\mathrm{W}}(\mathbf{u},\hat{\mathbf{u}})
=
\frac{1}{T}
\sum_{t=0}^{T-1}
\left|
\operatorname{Norm}_{\Delta}
\!\left(
\hat{\mathbf{u}}^{\,t+1}
-
\mathbf{u}^{t}
\right)
-
\operatorname{Norm}_{\Delta}
\!\left(
\mathbf{u}^{t+1}
-
\mathbf{u}^{t}
\right)
\right|_{1}.
\]
The normalization is performed with the global mean and standard deviation of the transition differences.

\section{Knowledge Distillation Details} \label{sec:kd_details}
In this section, we provide additional results on the KD techniques used in this paper.

\paragraph{TREX} For TREX, we use 100 rollout steps starting from the initial condition by default. The noise is applied after 10 steps, with a scaling factor of $\sigma=1$. The TREX loss weight is set to 1 as well. The noisy rollouts are regenerated every 1000 epochs. 

\paragraph{IC-KD}
For the IC-KD baseline, we use a buffer of teacher-generated trajectories that we regenerate every 10 epochs, with the size of the buffer determined by the number of ground truth trajectories available. The teacher-generated rollouts are then also divided into sub-trajectories for n-step training.  Note that for the Brownian bridge task, the initial conditions are computationally intensive to compute because they are evolved from an earlier state.
\[
\mathcal{L}_{\mathrm{IC}}
=
\mathbb{E}_{\sol_0 \sim p_0}
\left[
\frac{1}{T}
\sum_{t=0}^{T-1}
\ell\left(
\student(\sol^t),\teacher(\sol^t)
\right)
\right],
\qquad
 \sol^{t+1}=\teacher(\sol^t)
\]

\paragraph{Relational KD} Relational KD \citep{park2019relational} aims to make the distance between an input pair $i, j$ from a mini-batch in the student feature space close to the distance in the feature space:
\begin{equation}
    \mathcal{L}_{\text{RKD}} =
    \sum_{i < j}
    \ell_{\delta} \Big(
    d(\mathbf{t}_i, \mathbf{t}_j),
    d(\mathbf{s}_i, \mathbf{s}_j)
    \Big),
\end{equation}
where $\mathbf{t}$ and $\mathbf{s}$ are the teacher and student features, respectively. The  loss is the Huber distance
\begin{equation}
    \ell_\delta(x, y) =
    \begin{cases}
    \frac{1}{2}(x - y)^2 & \text{if } |x - y| \le 1 \\
    |x - y| - \frac{1}{2} & \text{otherwise}
    \end{cases}.
\end{equation}

Since the feature spaces of the spatial neural network are very large, we use Gaussian sketching, which uses random Gaussian matrices $R$ to reduce the input features to a lower dimension for the teacher
\begin{equation}
\mathbf{t}_i = \mathbf{R}_T \tilde{\mathbf{t}}_i,
\qquad
(\mathbf{R}_T)_{ab} \sim \mathcal{N}\!\left(0,\tfrac{1}{k}\right),
\qquad
\mathbf{R}_T \in \mathbb{R}^{k \times d_T},
\end{equation}
and equivalently, the student model
\begin{equation}
\mathbf{s}_i = \mathbf{R}_S \tilde{\mathbf{s}}_i,
\qquad
(\mathbf{R}_S)_{ab} \sim \mathcal{N}\!\left(0,\tfrac{1}{k}\right),
\qquad
\mathbf{R}_S \in \mathbb{R}^{k \times d_S}.
\end{equation}
It can be shown that the error in the distances $ d(\mathbf{t}_i, \mathbf{t}_j)$ introduced by the Gaussian sketch is bounded \citep{holzmuller2023framework,johnson1984extensions}. We use $k=512$ in all experiments

The loss  $ \mathcal{L}_{\text{RKD}}$ is added to the main, ground truth loss with a scaling factor of $0.001$. We use the last-layer features for both the student and the teacher.

\section{Further Results}
\begin{figure}
    \centering
    \includegraphics[width=1\linewidth]{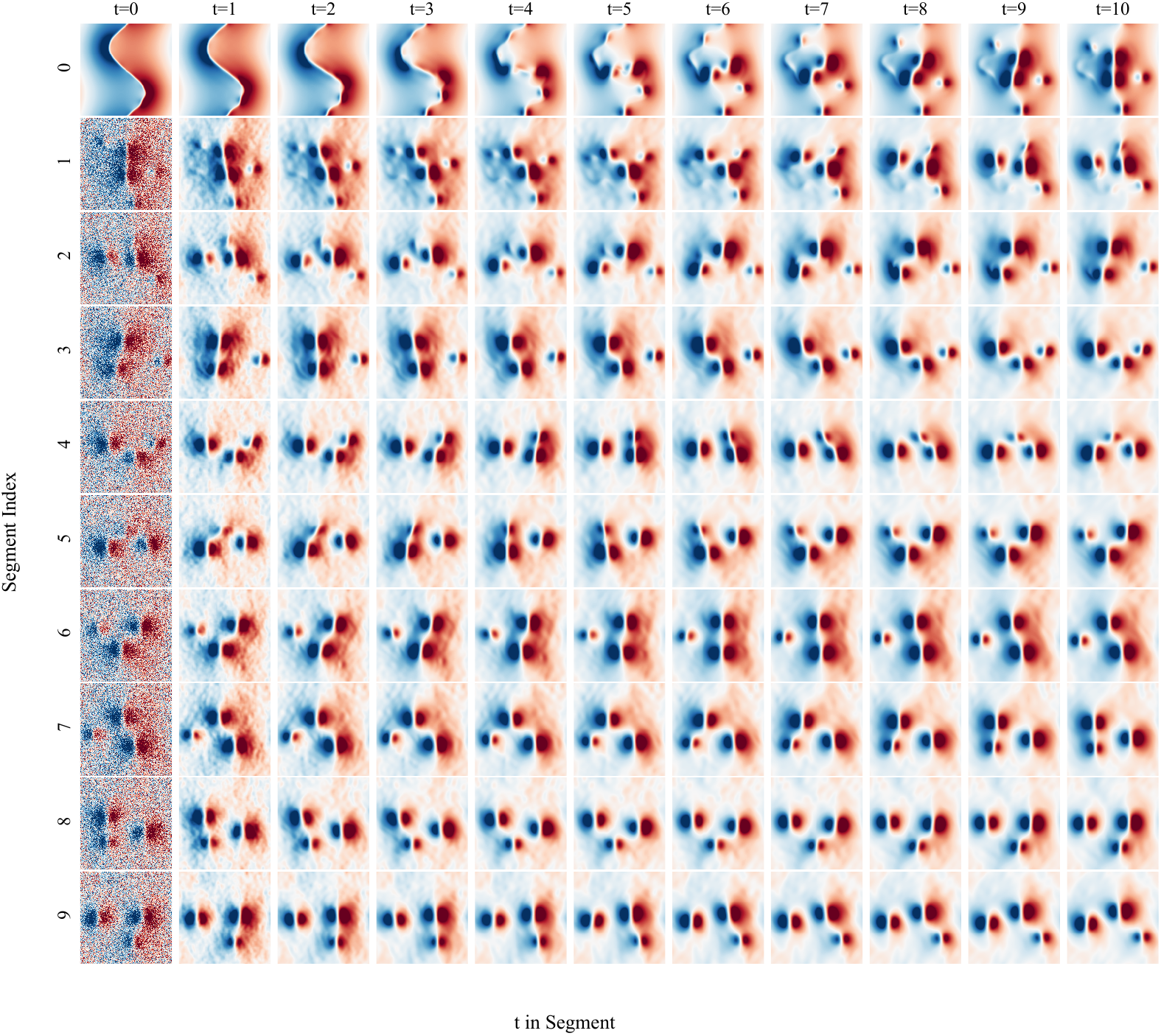}
    \caption{Example trajectory (velocity-x channel) generated using the teacher for $\sigma=1$ on NS-SVS. After each sub-trajectory (row), the noise is applied.}
    \label{fig:noisyrolloutsvs}
\end{figure}

\begin{figure}
    \centering
    \includegraphics[width=1\linewidth]{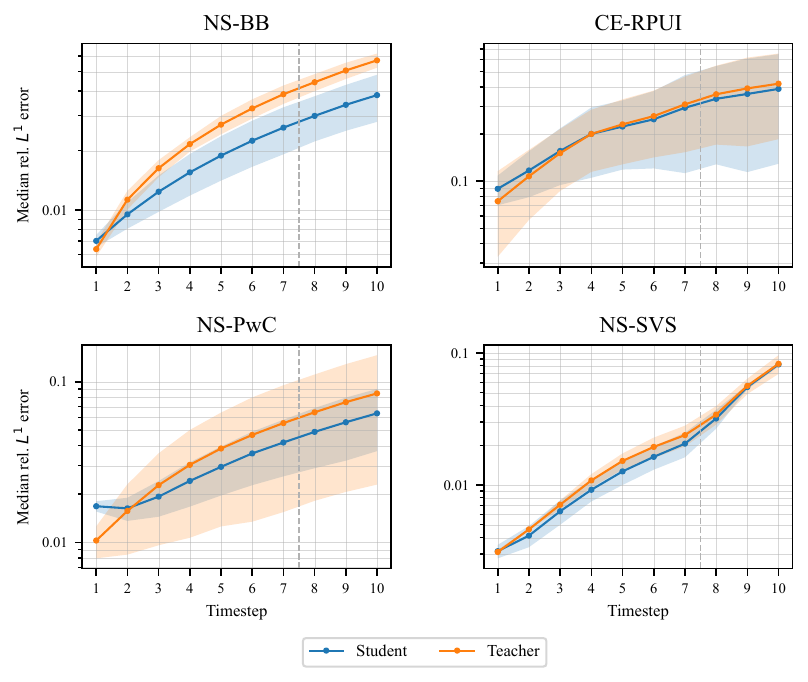}
    \caption{Error over time for the student trained with TREX (8 trajectories). The shaded area shows the 95\% confidence interval of the mean (3 seeds). The dashed line shows the time horizon seen during training.}
    \label{fig:all_error_growth}
\end{figure}

\cref{tab:main-results} shows the main results for the median relative $L^1$ metric with the 95\% confidence interval of the mean. For completeness, we also show the relative $L^1$ error for all time steps (\cref{tab:main-results-all}). Additionally, we report the training durations in \cref{tab:training_time}. \cref{fig:noisyrolloutsvs} shows a full example rollout for the NS-SVS task and \cref{fig:all_error_growth} shows the error growth for all 4 considered tasks. 

\sisetup{table-format=1.3(3),detect-weight=true,detect-inline-weight=math}
\begin{table}[h]
\centering
\caption{Median rel. $L^1$ metric of the \textit{last} time step. Entries show the mean with its 95\% confidence interval calculated over three seeds.}
\label{tab:main-results}
\begin{tabular}{lSSSS}
\hline
 {Method} & {$N=4$} & {$N=8$} & {$N=16$} & {$N=32$}  \\
\hline
\multicolumn{5}{c}{{CE-RPUI}} \\
\hline
 {TREX} & \bfseries 0.421(15) & \bfseries 0.295(182) & \bfseries 0.197(38) & 0.144(10)  \\
 {IC-KD} & 0.440(77) & 0.304(167) & 0.223(105) & 0.145(6)  \\
 {Relational KD} & 0.608(99) & 0.446(68) & 0.296(31) & 0.196(20)  \\
 {Only GT} & 0.608(86) & 0.458(92) & 0.296(11) & 0.199(18)  \\
\midrule
 {Teacher} & 0.446(74) & 0.310(157) & 0.223(108) & 0.143(9)  \\
\hline
\multicolumn{5}{c}{{NS-SVS}} \\
\hline
 {TREX} & \bfseries 0.043(37) & \bfseries 0.021(5) & \bfseries 0.014(1) & \bfseries 0.011(2)  \\
 {IC-KD} & 0.045(32) & 0.025(5) & 0.018(3) & 0.015(4)  \\
 {Relational KD} & 0.178(49) & 0.050(20) & 0.021(4) & 0.011(1)  \\
 {Only GT} & 0.171(53) & 0.060(21) & 0.023(5) & 0.011(1)  \\
\midrule
 {Teacher} & 0.043(30) & 0.024(5) & 0.018(5) & 0.015(3)  \\
\hline
\multicolumn{5}{c}{{NS-PwC}} \\
\hline
 {TREX} & \bfseries 0.056(3) & \bfseries 0.042(17) & 0.032(12) & 0.021(3)  \\
 {IC-KD} & 0.067(12) & 0.053(39) & 0.042(38) & 0.029(7)  \\
 {Relational KD} & 0.175(14) & 0.059(11) & \bfseries 0.027(3) & 0.013(1)  \\
 {Only GT} & 0.282(81) & 0.068(11) & 0.029(2) & \bfseries 0.013(1)  \\
\midrule
 {Teacher} & 0.070(11) & 0.055(40) & 0.044(35) & 0.031(5)  \\
\hline
\multicolumn{5}{c}{{NS-BB}} \\
\hline
 {TREX} & \bfseries 0.036(7) & \bfseries 0.026(7) & 0.023(6) & 0.016(8)  \\
 {IC-KD} & 0.041(10) & 0.034(4) & 0.031(5) & 0.023(9)  \\
 {Relational KD} & 0.111(22) & 0.034(7) & \bfseries 0.015(1) & \bfseries 0.007(2)  \\
 {Only GT} & 0.178(77) & 0.041(15) & 0.015(3) & 0.007(1)  \\
\midrule
 {Teacher} & 0.046(10) & 0.039(5) & 0.033(3) & 0.025(8)  \\
\hline
\end{tabular}
\end{table}

\begin{table}[t]
\centering
\caption{Median rel. $L^1$ metric over \textit{all} timesteps. Entries show the mean with its 95\% confidence interval calculated over three seeds.}
\label{tab:main-results-all}
\begin{tabular}{lSSSS}
\hline
 {Method} & {$N=4$} & {$N=8$} & {$N=16$} & {$N=32$}  \\
\hline
\multicolumn{5}{c}{{CE-RPUI}} \\
\hline
 {TREX} & \bfseries 0.277(27) & \bfseries 0.182(65) & \bfseries 0.127(19) & 0.095(8)  \\
 {IC-KD} & 0.281(35) & 0.184(80) & 0.135(55) & 0.093(5)  \\
 {Relational KD} & 0.471(74) & 0.316(42) & 0.197(25) & 0.129(14)  \\
 {Only GT} & 0.466(67) & 0.327(49) & 0.201(11) & 0.130(5)  \\
\midrule
 {Teacher} & 0.282(38) & 0.185(78) & 0.129(55) & 0.084(5)  \\
\hline
\multicolumn{5}{c}{{NS-SVS}} \\
\hline
 {TREX} & 0.020(14) & \bfseries 0.009(2) & \bfseries 0.006(1) & 0.004(1)  \\
 {IC-KD} & 0.020(9) & 0.011(2) & 0.008(1) & 0.006(1)  \\
 {Relational KD} & 0.070(23) & 0.019(9) & 0.008(2) & 0.004(1)  \\
 {Only GT} & 0.068(24) & 0.022(9) & 0.008(2) & \bfseries 0.004(1)  \\
\midrule
 {Teacher} & 0.020(8) & 0.011(2) & 0.008(1) & 0.006(1)  \\
\hline
\multicolumn{5}{c}{{NS-PwC}} \\
\hline
 {TREX} & \bfseries 0.032(2) & \bfseries 0.024(7) & 0.019(4) & 0.014(2)  \\
 {IC-KD} & 0.037(9) & 0.028(16) & 0.023(14) & 0.017(3)  \\
 {Relational KD} & 0.105(11) & 0.037(7) & \bfseries 0.018(2) & 0.010(1)  \\
 {Only GT} & 0.172(48) & 0.043(8) & 0.019(2) & \bfseries 0.010(1)  \\
\midrule
 {Teacher} & 0.039(11) & 0.029(16) & 0.024(14) & 0.018(2)  \\
\hline
\multicolumn{5}{c}{{NS-BB}} \\
\hline
 {TREX} & \bfseries 0.022(5) & \bfseries 0.016(5) & 0.013(2) & 0.009(3)  \\
 {IC-KD} & 0.023(6) & 0.019(4) & 0.017(1) & 0.012(4)  \\
 {Relational KD} & 0.067(12) & 0.022(4) & \bfseries 0.010(1) & \bfseries 0.005(1)  \\
 {Only GT} & 0.108(50) & 0.026(9) & 0.010(2) & 0.005(1)  \\
\midrule
 {Teacher} & 0.026(7) & 0.022(2) & 0.018(1) & 0.014(4)  \\
\hline
\end{tabular}
\end{table}
\begin{table}[t]
  \centering

  \caption{Average training duration in minutes on NS-PwC.}
  
  \sisetup{table-format=3.1(2),detect-weight=true,detect-inline-weight=math}
  \begin{tabular}{lSSSS}
    \toprule
    Method & {$N=4$} & {$N=8$} & {$N=16$} & {$N=32$} \\
    \midrule
    TREX & 47.1 \pm 1.9 & 99.8 \pm 3.2 & 147.4 \pm 1.4 & 291.4 \pm 4.6 \\
    IC-KD & 85.2 \pm 2.3 & 117.7 \pm 4.5 & 193.5 \pm 1.4 & 338.7 \pm 6.1 \\
    Relational KD & 38.2 \pm 1.4 & 72.9 \pm 2.7 & 143.1 \pm 3.8 & 280.2 \pm 9.8 \\
    Only Ground Truth & 37.3 \pm 0.4 & 70.2 \pm 0.4 & 138.4 \pm 4.3 & 271.5 \pm 7.8 \\
    \bottomrule
  \end{tabular}
  \label{tab:training_time}
\end{table}

\FloatBarrier
\subsection{Self-Distillation}
We also perform a self-distillation experiment to further examine the influence of the teacher pretraining. We use the TFNO trained only on ground-truth samples as the teacher for distillation into a freshly initialized model of the same architecture. The results in \cref{tab:selfdist} show that, while self-distillation has a small positive regularizing effect, the main advantage comes from distilling the pretrained, higher-capacity foundation model.

\begin{table}[h]
    \centering
    \caption{Self-distillation experiment using TREX on CE-RPUI (4 trajectories). }
    \begin{tabular}{cc}
    \toprule
      Method  & Median rel. $L^1$ \\
    \midrule
     Only GT    & 0.608\\
     TREX  (self-distillation) & 0.585\\
     TREX  (from Poseidon) & 0.421\\
     \bottomrule
    \end{tabular}
    \label{tab:selfdist}
\end{table}

\FloatBarrier
\subsection{Equivariance} 
\begin{figure}
    \centering
    \includegraphics[width=1\linewidth]{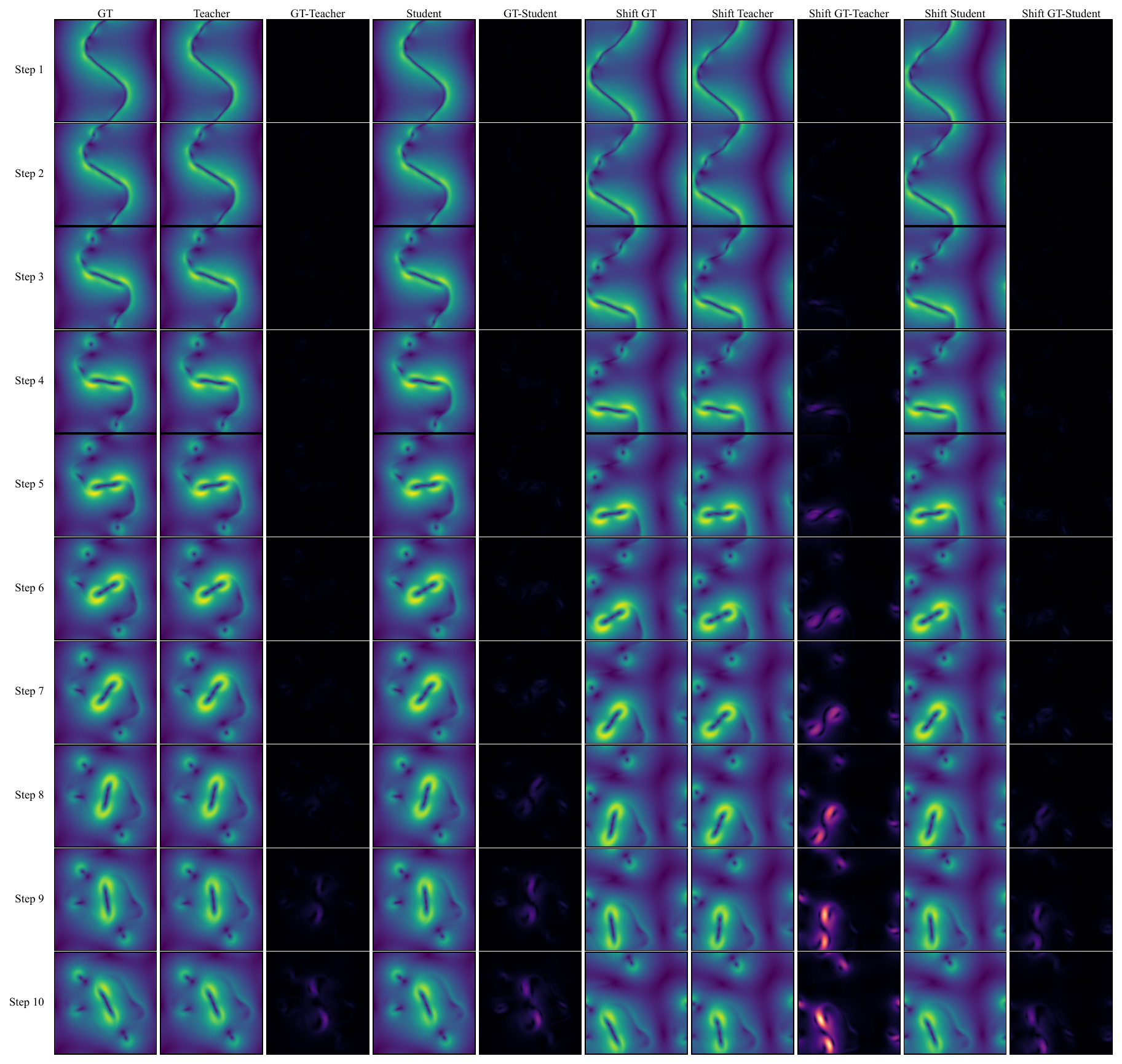}
    \caption{Full shifted-rollout comparison on NS-SVS. Each row shows one autoregressive step. The teacher accumulates larger errors under the spatial shift, while the equivariant student remains consistent with the shifted ground truth throughout the rollout.}
    \label{fig:apx-equi}
\end{figure}

As a complement to the final-step comparison in \cref{fig:equi-svs-reduced}, \cref{fig:apx-equi} shows the complete shifted rollout on NS-SVS. On the unshifted rollout, the teacher initially appears slightly closer to the ground truth than the student. After applying the spatial shift, however, the teacher prediction quickly degrades, starting around steps 3/4, while the student remains consistent throughout the rollout.

\end{document}